\documentclass[letterpaper, 10 pt, conference]{ieeeconf}  

\IEEEoverridecommandlockouts                              

\usepackage{graphicx} 
\usepackage{amsmath} 
\DeclareMathOperator*{\argmin}{arg\,min}
\usepackage{microtype}

\graphicspath{images}

\usepackage{multirow}

\usepackage{fancyhdr}
\title{\LARGE \bf
Fast, Compact and Highly Scalable Visual Place Recognition through Sequence-based Matching of Overloaded Representations
}

\author{Sourav Garg and Michael Milford
\thanks{SG and MM are with the QUT Centre for Robotics, School of Electrical Engineering and Robotics at the Queensland University of Technology. This work received funding from AOARD grant FA2386-19-1-4079.}
\thanks{Email: {\tt\small s.garg@qut.edu.au}}%
}

\begin{document}

\maketitle
\thispagestyle{fancy}
\pagestyle{plain}

\begin{abstract}

Visual place recognition algorithms trade off three key characteristics: their storage footprint, their computational requirements, and their resultant performance, often expressed in terms of recall rate. Significant prior work has investigated highly compact place representations, sub-linear computational scaling and sub-linear storage scaling techniques, but have always involved a significant compromise in one or more of these regards, and have only been demonstrated on relatively small datasets. In this paper we present a novel place recognition system which enables for the first time the combination of ultra-compact place representations, near sub-linear storage scaling and extremely lightweight compute requirements. Our approach exploits the inherently sequential nature of much spatial data in the robotics domain and inverts the typical target criteria, through intentionally coarse scalar quantization-based hashing that leads to more collisions but is resolved by sequence-based matching. For the first time, we show how effective place recognition rates can be achieved on a new very large 10 million place dataset, requiring only 8 bytes of storage per place and 37K unitary operations to achieve over 50\% recall for matching a sequence of 100 frames, where a conventional state-of-the-art approach both consumes 1300 times more compute and fails catastrophically. We present analysis investigating the effectiveness of our hashing overload approach under varying sizes of quantized vector length, comparison of near miss matches with the actual match selections and characterise the effect of variance re-scaling of data on quantization. Resource link: \tt{https://github.com/oravus/CoarseHash}

\end{abstract}

\section{INTRODUCTION}
Visual Place Recognition (VPR) is a key capability for a mobile robot, enabling it to localize itself within a known environment. The topic has been extensively researched for decades~\cite{lowry2016visual} with researchers exploring different aspects of the problem, such as dealing with appearance~\cite{milford2012seqslam,naseer2018robust} and viewpoint variations~\cite{garg2018lost,gawel2018x}, and large-scale localization~\cite{cummins2008fab,doan2019scalable,le2019btel} and navigation~\cite{chancan2020mvp}. FAB-MAP~\cite{cummins2008fab} was one of the earliest VPR methods to demonstrate large-scale mapping, also incorporated into visual SLAM systems like LSD-SLAM~\cite{engel2014lsd}. Large-scale retrieval has also been a topic of significant interest in the computer vision community, leading to highly-scalable retrieval techniques like BoVW~\cite{sivic2003video} and VLAD~\cite{jegou2010aggregating}.

Such VPR and retrieval solutions are typically characterized by their ability to compactly represent places for low overall storage and linear growth in terms of time and memory requirements during deployment. These algorithms typically trade off storage footprint with computational requirements or vice versa to achieve high performance. A vast literature exists for developing compact place representations~\cite{lowry2018lightweight,arroyo2016fusion} and demonstrating sub-linear scaling in both computation time~\cite{gionis1999similarity,vysotska2017relocalization} and recently storage~\cite{yu2018rhythmic,le2019btel} requirements.  

\begin{figure}
    \centering
    \includegraphics[scale=0.33,trim={0 6.75cm 4.5cm 0},clip]{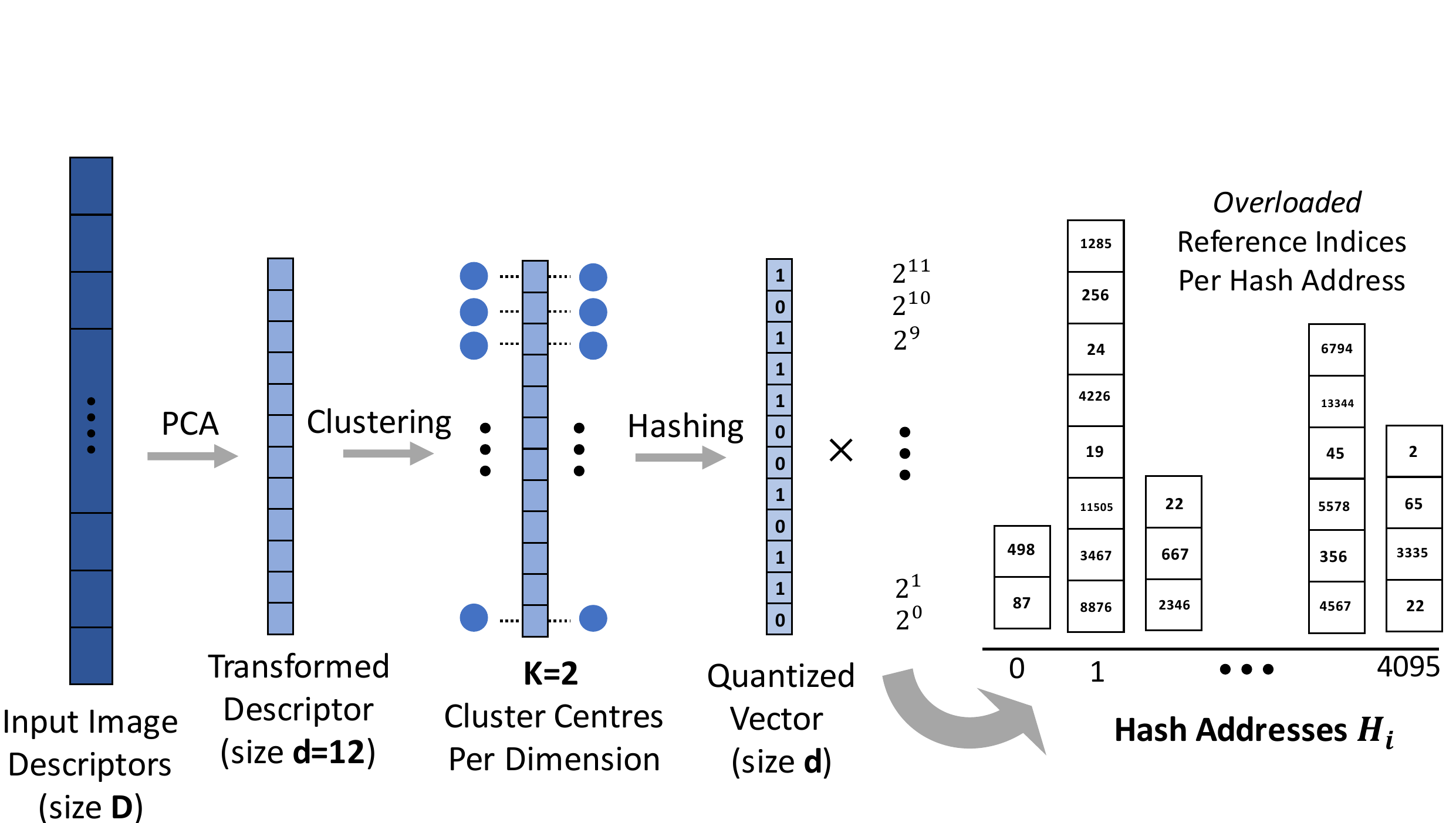}
    \caption{\textit{Coarse Hash}: Our hashing approach based on coarse scalar quantization generates short binary vectors (hash addresses) with long lists of inverted indices of reference data. When querying the hash space, collisions due to overloaded lists are resolved using sequence-based matching.}
    \label{fig:schematic}
\end{figure}

In this paper, we present a novel visual place recognition solution that 1) generates and uses highly compact place representations leading to very small storage footprint, 2) benefits from extremely fast retrieval, 3) achieves near sub-linear growth in storage, and 4) has the capability to perform real-time localization in a map containing 10 million places with very lightweight computational requirements (37 kflops for 1 Hz performance) - the largest \textit{sequential-nature} VPR benchmark to date. This unprecedented performance is achieved through massively overloading existing quantization and hashing techniques (ignoring the typical criteria used to optimize such techniques, see Figure~\ref{fig:schematic}), and then leveraging the sequential nature of robotic or autonomous vehicle data streams to disambiguate the resulting highly noisy single-frame matching performance.

Existing quantization methods have mainly focused on large vocabularies and longer resultant vectors to reduce the quantization error~\cite{jegou2010product,ge2013optimized}. This leads to better search accuracy and reduces the computation time requirements due to shorter lists of candidate matches which can then be re-ranked by superior techniques to find the best match~\cite{jegou2010product,babenko2016efficient}. This approach is highly suitable for retrieval systems where the searched database and the query data are un-ordered. However, for data of a spatio-temporal nature, for example, in VPR and localization, a different approach to quantization can be used to leverage the sequential nature of the information. In particular, we invert the target criteria for quantization techniques and deliberately aim for coarse quantization leading to much shorter vectors and longer lists of candidate matches. The typical disadvantage of long lists, a drastically increased probability of selecting a wrong match, is mitigated by the additional source of information in the form of temporal sequences, as also demonstrated conceptually in the original SeqSLAM~\cite{milford2012seqslam}. Where SeqSLAM and follow on work leveraged sequences to disambiguate very challenging perceptual data, here we take the complementary approach of deliberately degrading matching performance to gain significant advantages in compute and storage requirements. This approach enables successful place recognition performance with extremely compact representations. 

\section{Literature Review}
\label{sec:LitRev}
From large-scale place recognition and mapping solutions~\cite{paul2010fab,kejriwal2016high} based on traditional feature representations~\cite{lowe2004distinctive,bay2008speeded} to creating compact representations~\cite{lowry2018lightweight,arroyo2016fusion,chancan2020hybrid} and sequential-matching pipelines~\cite{vysotska2017relocalization} based on deep-learnt image descriptors~\cite{arandjelovic2016netvlad,chen2014convolutional}, the visual place recognition literature has grown rapidly in recent years. 

For large-scale image retrieval and place recognition, apart from retrieval techniques based on data structures like trees~\cite{lejsek2011nv,liu2005investigation,hou2017tree,schlegel2018hbst} and graphs~\cite{dong2011efficient,wang2012scalable,harwood2016fanng,gollub2017partitioned,garcia2017hierarchical}, quantization and hashing based methods have also been well explored. Vector quantization~\cite{gersho2012vector}, Product Quantization~\cite{jegou2010product}, Scalar Quantization~\cite{brandt2010transform,jegou2008hamming}, and their improved versions~\cite{ge2013optimized,sandhawalia2010searching,kalantidis2014locally} have been demonstrated to be highly scalable in terms of computation time. Similarly, hashing~\cite{gionis1999similarity,andoni2006near,vysotska2017relocalization,weiss2009spectral,han2017mild} and efficient indexing~\cite{jegou2010product,babenko2016efficient,babenko2014inverted,cieslewski2017efficient} techniques have been shown to be capable of retrieving accurate matches in nearly constant time. However, an overall low storage footprint is not always guaranteed with these methods. Furthermore, the aim for most of these methods is to either reduce the quantization error or to retrieve a short list of candidate matches for effectively finding the best match. With additional sequential information, some of the conditions of these existing systems can be relaxed to allow longer lists of candidate matches, which can then be filtered by well-established sequence-search techniques for VPR~\cite{milford2012seqslam,vysotska2016lazy,pepperell2016routed}.

Within the context of VPR, some of the existing works that leverage sequential information include the use of sequential cyclic patterns~\cite{yu2018rhythmic} and binary tree encoding approach to directly infer the matched index~\cite{le2019btel}. Although achieving sub-linear growth in storage, these methods have their limitations due to certain assumptions: the former expects particular frequency patterns to occur within the encoded traverses and the latter requires the entire encoded traverse to maintain the order of adjacency. In our proposed system, we do not make any such assumptions about the data, instead we rely on unsupervised data transformation~\cite{brandt2010transform} that suits the following scalar quantization-based encoding~\cite{jegou2008hamming,sandhawalia2010searching}.

\section{Proposed Approach}
\label{sec:PA}
The existing quantization and hashing techniques for large-scale retrieval mainly focus on unordered datasets~\cite{jegou2010product,ge2013optimized,brandt2010transform}. In the context of visual place recognition and localization, the underlying datasets are usually sequentially ordered~\cite{milford2012seqslam,cummins2008fab,maddern20171}. Our proposed VPR pipeline demonstrates the effective use of hashing and inverted-index lists for sequential data. In particular, we show that using coarse scalar-quantization based hashing~\cite{brandt2010transform}, one can allow a large number of collisions which can then be resolved by sequence-based matching~\cite{milford2012seqslam}. This enables storage-efficient encoding of the image descriptors with much shorter codes and correspondingly long inverted lists of candidate matches. This is in contrast to the existing state-of-the-art large-scale retrieval methods where relatively longer codes and short lists are often the preferred choices to achieve high performance~\cite{jegou2010product,babenko2014inverted}. In the following, we first describe the quantization process for the reference and the query datasets and then present the sequence-based matching approach.

\subsection{Reference Data to Hash Addresses}
The reference data for VPR, available beforehand in online operations, is quantized and hashed to an integer address space where each hash address is linked to multiple reference image indices. This is obtained as described below:

\subsubsection{Image to Descriptor}
The raw image data of size $N_x$ is first converted into $D$-dimensional global image representations using a state-of-the-art image description technique~\cite{arandjelovic2016netvlad,torii201524,garg19Semantic}. We used NetVLAD~\cite{arandjelovic2016netvlad} for this purpose, however, any other method can be used as a drop-in replacement. 

\subsubsection{PCA Transformation}
The $N_x \times D$ feature matrix obtained above is then transformed into a decorrelated orthogonal space using PCA. We use Incremental PCA~\cite{ross2008incremental,levey2000sequential,scikitLearn} for this purpose to keep the transformation computationally tractable. We only retain the first $d$ components of the transformed feature matrix.

\subsubsection{Hashing}
The transformed feature matrix $N_x \times d$, being decorrelated and mutually-orthogonal, can be independently quantized along each of its dimensions~\cite{jegou2008hamming,brandt2010transform}. This scalar quantization per dimension can be obtained using $K$-means clustering, $K$ being the number of quantization bins. While we keep $K$ fixed for every dimension, we note that some  existing approaches use different values of $K$ depending on the variance along that dimension~\cite{brandt2010transform,weiss2009spectral,sandhawalia2010searching}.

\begin{equation}
    q_{x_{ij}} = \argmin_{k \in [1,K]} \left|x_{ij} - c_{kj}\right| \quad \forall j \in [1,d]
    \label{eq:quantRef}
\end{equation}
where $x_i$ and $q_{x_i}$ represent the transformed reference image descriptor and its quantization index vector respectively and $c_k$ represents the $k$th cluster center along the $j$th dimension. 

The quantization index vectors are converted into integer hash addresses as below:
\begin{equation}
    h_{x_i} = \sum_{j=1}^d K^{(d-j)}  q_{x_{ij}}
    \label{eq:hashRef}
\end{equation}

At each of these hash addresses, multiple image indices are stored as a list. The maximum number of addresses possible are $K^d$, however, in practice, a number of these addresses comprise empty lists due to collisions at other hash addresses. 

\subsection{Query Searching}

Query images are first converted into global representations using the same description method as for the reference data. The transformation matrix obtained from PCA training of the reference data is used for transforming the query descriptors to $d$-dimensional vectors. A hash address for a given query vector is obtained using Equation~\ref{eq:quantRef} and \ref{eq:hashRef}. Therefore, a list of matched reference indices for a given query image can be obtained with the computational complexity of $\mathcal{O}(1)$. The list of matched indices represents the candidate matches for the sequence-based filtering described in the subsequent section. For single frame-based matching, we only store the best match corresponding to a hash address instead of a list of reference indices. This is obtained on the basis of minimum quantization error:

\begin{equation}
    i_{single} = \argmin_{i \in [1,N_x]} \enskip \sum_{j=1}^d\min_k\left| x_{ij}-c_{kj}\right|
    \label{eq:singleBest}
\end{equation}
\noindent which can be computed along with the hashing of reference data before the query phase begins. As no further computation is needed during the query search, search complexity of $\mathcal{O}(1)$ is retained. We highlight the effect of the above selection procedure on single frame-based matching performance in Section~\ref{sec:Results}.

\subsubsection*{Queries Landing at Unoccupied Hash Addresses}
Due to perceptual aliasing, images (and their corresponding descriptors) obtained from revisited places are not guaranteed to exactly match to their ground truth in the reference database. Therefore, a quantization index vector of a query image can lead to a hash address which is not associated to any reference image index (or a list of indices). Such queries are assigned the nearest occupied hash address (numerically closest) from the sorted list of all the occupied addresses. This search is completed in $\mathcal{O}(\log_2H_o)$ time where $H_o$ is the number of occupied addresses. Alternatively, it would also be possible to pre-compute and store the nearest neighbours in the hash address space during the training, representing a different operating location in the trade-off between storage footprint and query time.

\subsection{Sequence-based Matching}
Inspired by SeqSLAM~\cite{milford2012seqslam}, we use a similar strategy of disambiguating place matches using sequential information. A combined set of reference candidates obtained from the lists of potential matches of a query sequence are probed to find the best match. For this purpose, the distance between a quantized query and reference vector is defined as below:
\begin{equation}
    \delta(x_i,y_t) = \sum_{j=1}^d\left|c_{jq_{x_{ij}}} - c_{jq_{y_{tj}}}\right|
    \label{eq:distCont}
\end{equation}
where $q_{x_{ij}}$ and $q_{y_{tj}}$ refer to the cluster centers assigned to the $j$th dimension of the reference and query vectors $x_i$ and $y_t$ respectively. $q_{x_i}$ for any reference image is directly obtained from the base $K$ representation of its hash address $h_{x_i}$ (see Equation~\ref{eq:hashRef}). The distance calculation in Equation~\ref{eq:distCont} is similar to the Symmetric Distance Computation (SDC) defined in~\cite{jegou2010product}.

For a given $L$-length sequence of quantized query vectors centered at $y_t$, the lists of matching reference indices are obtained from their respective hash addresses $h_{y_t}$. The set of unique reference indices, $r$, obtained from these lists is then used for sequence searching and the best match is obtained as below:

\begin{equation}
    i_{seq} = \argmin_{i\in[1,N_r]}\sum_{l=-L/2}^{L/2-1}\delta(r_{il},y_{tl})
    \label{eq:seqSearch}
\end{equation}
where $r_i$ represents a shortlisted candidate from a set of size $N_r$. Here, we assume a constant velocity between consecutive samples of reference and query data.

\section{Experimental Setup}
\label{sec:ExpSetup}
\subsection{Datasets}
We used two types of datasets in our experiments: a newly collected large-scale localization dataset - FAS100K and a commonly used benchmark dataset for large-scale image retrieval - Deep1B~\cite{babenko2016efficient}. The datasets chosen are a result of the sparsity of very large scale spatial navigation datasets: we describe pre-processing, benchmarking and analysis in depth to show that our treatment of the data is valid and appropriate.

\paragraph{FAS100K} This dataset is comprised of two traverses of $238$ and $130$ kms respectively where the latter is a partial repeat of the former. The data was collected using stereo cameras in Australia under sunny day conditions. It covers a variety of road and environment types including urban and rural areas. The raw image data from one of the cameras streaming at $5$ Hz constitutes $63650$ and $34497$ image frames for the two traverses respectively. We sub-sample these image sets with GPS information such that consecutive image frames are $5$ meters apart. The sub-sampled data of size $47781$ and $26638$ respectively form the reference and query traverses for our experiments. The images from both the traverses are converted into $4096$-dimensional global descriptors using the NetVLAD~\cite{arandjelovic2016netvlad} representation.

\paragraph{Deep1B} This is a recently introduced 1 billion image descriptor dataset~\cite{babenko2016efficient} comprising $96$-dimensional PCA-transformed descriptors. The original images or their hyperlinks are not publicly available, however, the dataset is one of the largest of its kind and is typically used for benchmarking large-scale nearest neighbor retrieval~\cite{babenko2016efficient,douze2018link,chiu2019learning,johnson2017billion}. Unlike localization datasets~\cite{maddern20171,milford2012seqslam,cummins2008fab}, these 1 billion descriptors are unordered and are temporally unrelated. In order to use this dataset in conjunction with FAS100K for our localization experiments, we perform pre-processing to create three new datasets of varying size: 20K, 1M, and 10M, comprising approximately $20,000$; $1,000,000$; and $10,000,000$ descriptors in both reference and query sets.

\paragraph{20K, 1M, and 10M} 
These three new `localization' datasets use FAS100K and Deep1B in parts (see Table~\ref{tab:datasetDescription}). For the reference traverse, 20K uses the first 10K samples from both Deep1B and FAS100K reference data (out of $47781$). 1M and 10M use the first 1 million and 10 million samples from Deep1B respectively and the entire reference data from FAS100K, leading to $1,047,781$ and $10,047,781$ reference descriptors. For the query counterparts of these reference datasets, the Deep1B part of the data in each of the three cases is re-used but with two different noise models of varying noise intensity, described later in this section. These Deep1B query datasets are then appended with FAS100K query data (out of $26638$): only the first 10k for the 20K dataset, and the entire query traverse for 1M and 10M datasets. Before concatenation, the source datasets are pre-processed as described in the following section.

\subsection{Data Pre-Processing and Concatenation}
\label{sec:DataPreProcess}
We perform the following pre-processing before concatenating the Deep1B and FAS100K datasets to obtain either of 20K, 1M, and 10M datasets:
    \paragraph{Homogenization} As also indicated earlier, the raw adjacent samples in the Deep1B descriptor data are unrelated, unlike typical localization datasets. In order to emulate local temporal perceptual similarity within the Deep1B dataset, we perform a sliding window average of the raw descriptors using a window size $w$ to make the data somewhat locally similar. $w$ is set to $40$ in our experiments which is equivalent to $200$ meters in the FAS100K spatial dataset. The homogenization enables the modified dataset to behave like a localization dataset in terms of matching performance in a region. Furthermore, this makes the Deep1B dataset more appropriately challenging: in its raw form, sequences of its unordered data are highly distinctive and can be easily matched.
    
\begin{figure}
    \centering
    \includegraphics[scale=0.5]{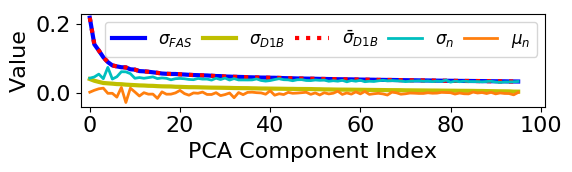}
    \caption{Standard deviation of PCA-transformed Deep1B descriptors $\sigma_{D1B}$ is modified to $\bar\sigma_{D1B}$ to match with PCA-transformed FAS100K descriptors $\sigma_{FAS}$ before concatenation to form the 20K dataset.}
    \label{fig:dataMergeStats}
\end{figure}

    \paragraph{Aligning Descriptor Dimensions and Variance} For the Deep1B dataset, neither the original images corresponding to its descriptors nor the PCA transformation matrix are publicly available. Hence, for any of the combined datasets (20K, 1M, and 10M), we concatenate the Deep1B and FAS100K datasets following a two step procedure. 1) PCA training is done independently on the reference data of Deep1B and FAS100K to match the descriptor length to $D=96$ and to obtain mutually decorrelated descriptor components ordered by their variance. 2) As the distribution of variance across the principal components for both the datasets is different, we re-scale the standard deviation of transformed Deep1B dataset to match it to FAS100K data before finally concatenating them. For the query data counterparts, in case of FAS100K, the query data is transformed using the PCA parameters of FAS100K training data, and in case of Deep1B, the noisy version (explained below) of the PCA-transformed and variance-equalized Deep1B data is used. Figure~\ref{fig:dataMergeStats} shows how the standard deviation for PCA-transformed Deep1B descriptors is matched to the PCA-transformed FAS100K descriptors.
    
    \paragraph{Noise Model for Query Data} We add noise to the modified Deep1B data to obtain a corresponding query dataset. The noise is added from a random normal distribution with mean $\mu_n$ and variance $\sigma_n^2$. The mean and variance are calculated from the FAS100K dataset which is first transformed using PCA to match the dimension size $D$ of the Deep1B dataset, that is, $96$:
    \begin{equation}
    \Delta_f = x_i^f-y_i^f, \enskip 
    \mu_n = \sum_i^{N_f}\frac{\Delta_f}{N_f}, \enskip 
    \sigma_n^2 = \sum_i^{N_f}\frac{(\Delta_f-\mu_n)^2}{N_f}
    \end{equation}
    where $x_i^f$ and $y_i^f$ are corresponding descriptor pairs from the FAS100K dataset and $N_f$ is $26638$. $x_i^f$, $y_i^f$, $\mu_n$, and $\sigma_n$ are all $96$-dimensional vectors. Figure~\ref{fig:dataMergeStats} shows the mean and standard deviation of noise compared against the standard deviation of merged datasets. We use two variants of the query data in our experiments: QM1 with distribution parameters $\mu_n$ and $\sigma_n$ and QM2 with parameters $\mu_n$ and $2\sigma_n$. This is done to study the impact of noise on the performance. Note that the noise is only added to the Deep1B (not the FAS100K which already comes with natural image variation) parts of the full query datasets.

\begin{table}[h]
    \centering
    \caption{New Localization Datasets}
    \scalebox{0.7}{
    \begin{tabular}{|c|c|c|c|c|}
    \hline
        \multirow{2}{*}{Dataset Size} & \multicolumn{2}{c|}{Reference Data} & \multicolumn{2}{c|}{Query Data} \\
        \cline{2-5}
        & Deep1B & FAS100K & Deep1B & FAS100K \\
        \hline
        20K & 10,000 & 10,000 & 10,000 & 10,000\\
        1M & 1,000,000 & 47781 &  1,000,00 & 26638 \\
        10M & 10,000,000 & 47781 & 10,000,00 & 26638 \\
    \hline
    \end{tabular}
    }
    \label{tab:datasetDescription}
\end{table}

\subsection{Baseline for Comparative Study}
We compare our proposed system with a baseline system which can be considered as a modified version of SeqSLAM~\cite{milford2012seqslam}. While both the systems are benchmarked on the same datasets: 10K, 1M, and 10M, both approaches are constrained to the same minimal storage footprint: in the case of the baseline, by limiting the number of descriptor dimensions. We perform PCA on the input dataset, similar to the process described in Section~\ref{sec:PA}, and only retain the initial principal component(s) for the baseline system - this enables a fair comparison under similar storage constraints as the first principal component divides the whole data with maximum variance.

A second constraint on the baseline system is imposed during the sequence matching process for equating the computation time of sequence searching. As in SeqSLAM, the baseline system linearly searches for matching candidates. However, we limit the total number of candidates to match with based on the average number of candidates shortlisted by our proposed approach $N_r$.

\subsection{Parameters Settings}
In our experiments conducted on the three datasets: 10K, 1M, and 10M, we use the following parameter settings: 1) The input descriptor dimensions for all the datasets is fixed to $D=96$; 2) the PCA-transformed descriptor dimension $d$ is set to $12$, $20$, and $24$ for the three datasets respectively for our proposed approach, while for the baseline system $d$ is set to $1$ to match the storage footprint; 3) Sequence length $L$ used for different experiments is $1$, $50$, and $100$ where $1$ implies single image based retrieval; 4) The number of cluster centers per dimension $K$ is fixed to $2$ for our proposed system for all the experiments. In order to keep the running time of experiments tractable, we only query a single frame or a sequence of frames every $z$th index. $z$ is set to $10$, $100$, and $10000$ for 20K, 1M, and 10M dataset respectively.

\subsection{Evaluation}
For benchmarking the two methods, we use recall rate which is often used for evaluating place recognition~\cite{garg19Semantic} and image retrieval systems~\cite{jegou2010product,babenko2016efficient}. Recall rate is defined as the ratio of correctly matched queries to the total number of queries within a given localization radius. The localization radius is varied from $0$ to $20$ frames which is equivalent to a maximum of $100$ meters for the $5$ meter frame separation in the FAS100K section of the datasets and half-window size $w/2$ for the Deep1B chunk of the datasets.

\section{Results}
\label{sec:Results}
\paragraph{Proposed vs Baseline}
Figure~\ref{fig:perf_allData} shows the performance comparison between our proposed system and the baseline. It can be observed that the baseline system performs very poorly under the same storage constraints as compared to our system, despite using different query noise models and varying sequence lengths. It can further be observed that more noise leads to a faster reduction in performance when scaling up to very large datasets. Similarly, the reduction in the performance of the baseline system scales up with the size of the datasets; even the use of a longer sequence length does not recover much performance. From the FAS100K part of the query data ($267$ queries) of the 20K dataset, the baseline was only able to retrieve $1$ successful match within a localization radius of $100$ meters whereas the proposed system correctly recalled $98$ matches.

\newcommand{\scaleOne}{0.2}

\begin{figure}
    \centering
    \begin{tabular}{ccc}
        \multicolumn{3}{c}{\includegraphics[scale=0.4,trim={0 1.65cm 0 1cm},clip]{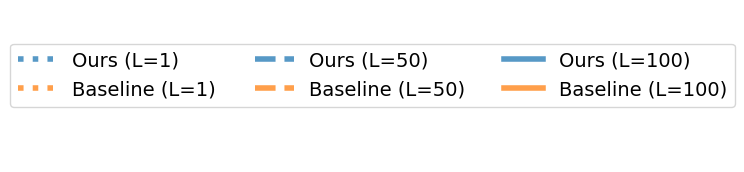}}\\
        \includegraphics[scale=\scaleOne]{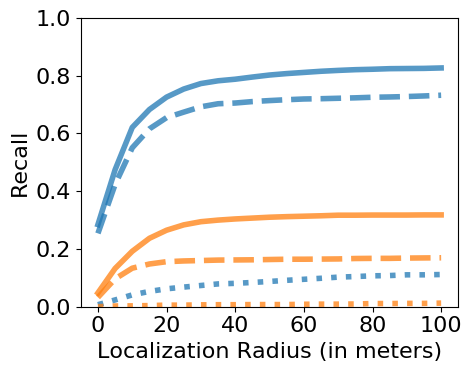} &
        \includegraphics[scale=\scaleOne]{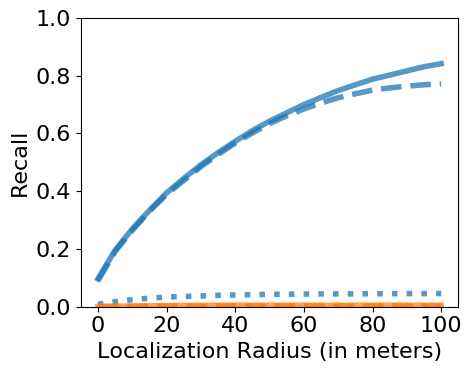} &
        \includegraphics[scale=\scaleOne]{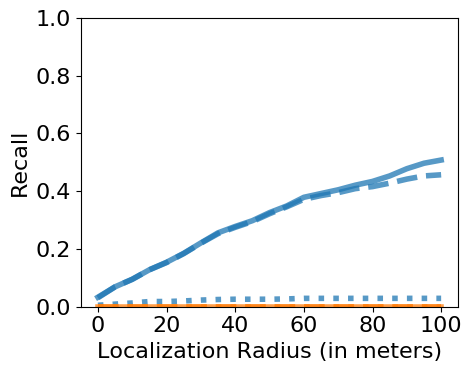} \\
        \includegraphics[scale=\scaleOne]{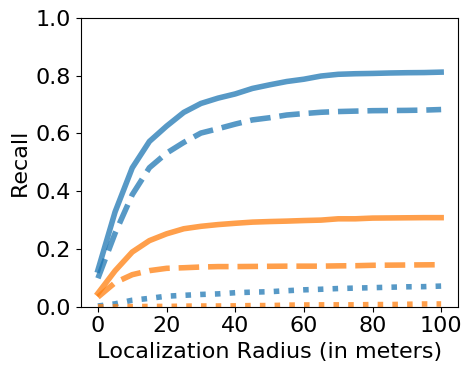} &
        \includegraphics[scale=\scaleOne]{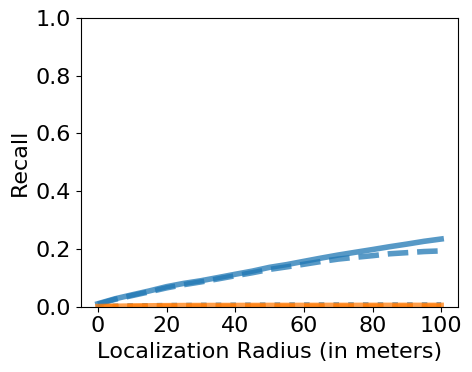} &
        \includegraphics[scale=\scaleOne]{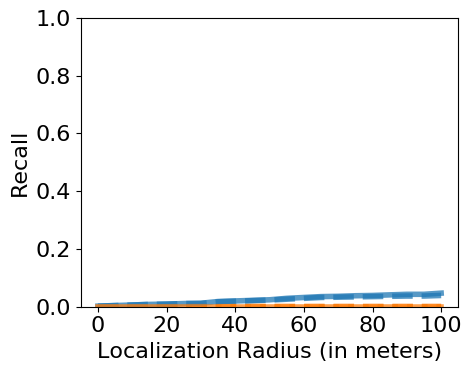} \\
        (a) 20K & (b) 1M & (c) 10M \\
    \end{tabular}
    \caption{Performance comparisons using two different noise models for the query data: QM1 (top) and QM2 (bottom). Line style represents different sequence lengths.}    
\label{fig:perf_allData}
\end{figure}

\newcommand{\scaleTwo}{0.33}

\begin{figure*}
    \centering
    \begin{tabular*}{\textwidth}{cccc}
    \includegraphics[scale=\scaleTwo]{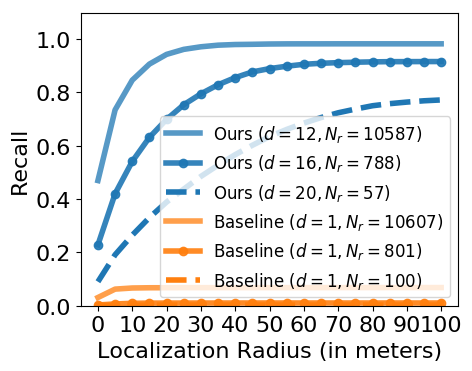}&
    \includegraphics[scale=\scaleTwo]{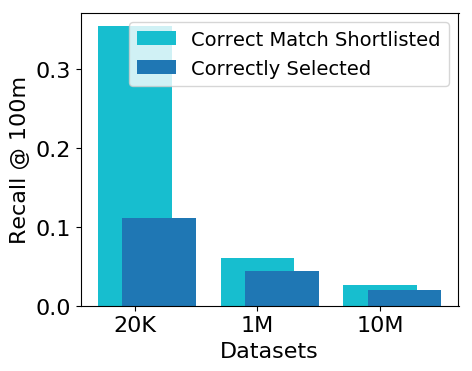}&
    \includegraphics[scale=\scaleTwo]{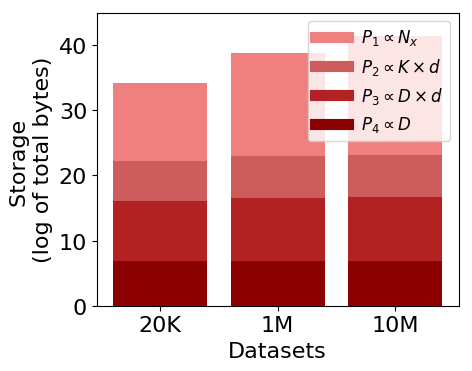}&
    \includegraphics[scale=\scaleTwo]{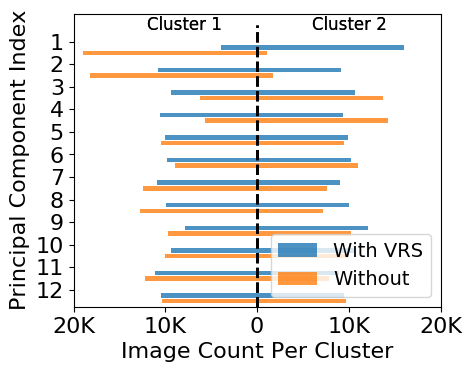}\\

    (a) & (b) & (c) & (d)
    \end{tabular*}
    \caption{(a) Performance variation for the 1M dataset with respect to vector lengths $d$. (b) Single Frame Matching: Performance comparison between the selected best match and the list of candidate matches using QM1. (c) Storage growth with respect to dataset size. (d) Cluster distribution of reference data (20K) across principal components of its binary vector both with and without variance re-scaling of the Deep1B dataset for unbiased dataset concatenation.}
\label{fig:otherResults}
\end{figure*}

\newcommand{\scaleThree}{0.07}

\paragraph{Longer Candidate Lists and Sequence Matching}
Figure~\ref{fig:otherResults}(a) shows the variation in performance when a relatively shorter vector length $d$ is chosen for 1M dataset. A short quantization vector leads to longer overloaded lists of matches $N_r$ which can then be effectively filtered by the sequence matching process. It can be observed that as $d$ decreases, $N_r$ increases and leads to high performance. However, the baseline system, despite the availability of more candidate matches, does not perform well.

\paragraph{Single Frame Matching - Best Match Selection}
For single frame matching, the selection of the best match out of the list of matches is based on the minimum quantization error as calculated using Equation~\ref{eq:singleBest}. In Figure~\ref{fig:otherResults}(b), we compare the percentage of correctly selected matches with the percentage of correct matches that existed \textit{somewhere} in the list of candidates, which can be regarded as a maximal upper bound on performance. This result indicates that the proposed system is generally able to achieve a higher recall when considering the correct matches within the list of candidates. The recall rate could be further improved beyond our match selection technique by using a re-ranking based on full descriptor matching~\cite{jegou2010product} or geometric verification~\cite{garg19Semantic}. Furthermore, with the increasing size of the dataset, the length of the quantized vector is also chosen to be proportionally longer. A longer quantization vector generally leads to a sparser distribution of reference indices with shorter lists of candidate matches. Therefore, the scope for further improving the single frame performance on very large datasets diminishes with the size of the dataset.

\paragraph{Storage Growth and Overall Footprint}
\label{subsec:storageGrowth}
Figure~\ref{fig:otherResults}(c) shows a comparison between the storage growth of different stored components of the proposed system, namely P1: Reference Indices to Hash Address Map, P2: Cluster Centers, P3: PCA Transformation Matrix, and P4: Mean Reference Descriptor. The absolute storage is represented as $\log_n$ of the raw values for visual clarity. It can be observed that P4 has a constant storage as it is proportional to the length of reference descriptors $D$; P2 and P3 grow sublinearly with the size of the database as they only depend on the choice of number of cluster centers $K$ and length of quantization vector $d$; P1 takes up the bulk of storage space as it is directly proportional to the the size of the reference database $N_x$. The overall storage used for the three datasets: 20K, 1M, and 10M for ours and the baseline system was $0.2$, $8.4$ and $80.4$ MB respectively.


\section{Discussion}
\label{sec:Disc}

\subsection{Computation Time Analysis}
Our proposed system is extremely fast as compared to the baseline system. The total number of unitary operations (additions and multiplications) required during query searching for both the systems are compared below:
    \subsubsection{PCA transformation} The subtraction of mean vector and matrix multiplication for PCA transformation require $D$ and $d(2D-1)$ operations respectively. $d$ for the baseline system is chosen to be smaller than that for the proposed system to match the overall storage footprint. As $d \ll D \ll N_x$, the computational advantage for baseline is minimal.
    \subsubsection{Quantization and Hashing} Quantizing a given query vector requires $d(2K-1)$ operations for assigning the cluster centers as defined in Equation~\ref{eq:quantRef}. Further, $2d-1$ operations are required to obtain a hash address from the quantized vector as defined in Equation~\ref{eq:hashRef}. These two steps are only required for the proposed system. However, these computations do not depend on the size of the database $N_x$ and are extremely fast in practice. 
    \subsubsection{Lists of Reference Candidates} For the proposed system, this is achieved by searching for the hash address key in the dictionary with values as the lists of reference indices. Hence, the computational complexity is $\mathcal{O}(1)$. For the baseline system, a query descriptor is compared against all the reference descriptors using Euclidean distance and requires $3dN_x$ operations. A list of candidates is then obtained by retaining $N_r$ candidates with lowest Euclidean distance. 
    \subsubsection{Sequence Matching} For a practical VPR scenario, sequence matching can be performed online requiring incremental computations for newly observed query images. Therefore, for the baseline and the proposed system, only $3N_r-1$ computations are required if cumulative sequence scores are stored for corresponding pairs of reference and query image indices. Unlike the baseline method, an exhaustive matching between query and reference data is not performed for the proposed system as it obtains a list of matching candidates from its hash address. Hence, for a paired sequence defined in Equation~\ref{eq:seqSearch} and the distance function defined in Equation~\ref{eq:distCont}, the proposed system requires $L'N_r(d/p+d-1)$ operations for sequence matching where $L'$ is the number of new pairs, $p$ is the number precision used which is $64$-bit for all the experiments, and $d/p$ represents the bitwise xor operation to find the distance in address space (Equation~\ref{eq:distCont}). As $L'<L \ll N_x$ and $N_r \ll N_x$, the computation time for this step for the proposed system is significantly smaller than the candidates' retrieval time for the baseline system described in the previous step which is $\mathcal{O}(N_x)$.
    
\subsection{Possible Throughput}
The compute time for our method primarily depends on $N_r$ which depends on the size of both the database ($N_x$) and the hash address space. The latter is given by $K^d$. With $K=2$, we fix $d$ so that it is just above $\log_2N_x$ which comes out to be $15$, $20$, and $24$ respectively for the 20K, 1M, and 10M datasets. In an ideal scenario with a uniform distribution of reference indices, there would be no more than $1$ reference index per hash address. However, in practice, we found the average number of reference indices ($N_r$) for a sequence of $50$ frames to be $74$, $57$, and $32$ for the three datasets, indicating a sub-linear growth in $N_r$ with respect to $N_x$. For the 10M dataset with $d=24$, $L'=L=50$, and $N_r=32$, sequence matching requires $37400$ unitary operations. On a hardware platform like a Jetson TX2, capable of $1.3$ TFLOPs~\cite{nvidiaTX2FLOPS}, the proposed method could potentially localize within the 10M reference database at a rate of 35 \textbf{MHz}. 

\subsection{Unbiased Dataset Concatenation}
Figure~\ref{fig:otherResults}(d) shows the distribution of reference indices across each of the dimensions (principal components on vertical axis) of the quantized vector for the 20K dataset. As we use two cluster centers per dimension, the left and the right side of the vertical dashed line represent clusters 1 and 2 respectively. Color indicates whether the variance for Deep1B dataset was re-scaled (blue) or not (orange), as described in Section~\ref{sec:DataPreProcess}. It can be observed that re-scaling of the variance leads to a uniform distribution of reference data across most of the dimensions. This indicates that concatenation of the Deep1B and FAS100K datasets does not favor any particular section of the 20K dataset. However, clustering is as expected imbalanced when no re-scaling is performed (orange) particularly for the first few components which leads to a disproportionate distribution of reference indices to hash addresses, and consequently poor performance.

\section{Conclusion}
In this paper, we have demonstrated a highly-scalable VPR pipeline that uses coarse scalar-quantization based hashing, leading to long lists of inverted reference indices due to shorter quantization vectors. The collisions in the hash space due to overloaded lists are then resolved by sequence-based matching. Our proposed system exhibits: low overall storage footprint, extremely fast retrieval, and near sub-linear storage growth with increasing size of the reference database, demonstrated on a new 10 million place dataset of \textit{sequential nature}.

\bibliography{main}
\bibliographystyle{IEEEtran}

\end{document}